\documentclass[letterpaper]{article} 
\usepackage{aaai2027} 
\makeatletter
\def\copyright@text{}
\makeatother

\usepackage[hyphens]{url} 
\usepackage{graphicx} 
\usepackage{natbib} 
\usepackage{caption} 
\usepackage{amsmath}
\usepackage{amssymb}
\usepackage{booktabs}
\usepackage{multirow}
\usepackage{array}

\usepackage{algorithm}
\usepackage{listings}
\usepackage{xcolor}
\usepackage[table]{xcolor}
\usepackage{subcaption}

\lstdefinestyle{pytorch}{
    language=Python,
    basicstyle=\ttfamily\scriptsize,
    commentstyle=\color{green!45!black},
    keywordstyle=\color{blue},
    stringstyle=\color{orange},
    numbers=none,
    frame=none,
    breaklines=true,
    columns=fullflexible,
    keepspaces=true,
    showstringspaces=false
}

\newcommand{\method}{PatchHead}

\title{PatchHead: Learning Spatial Patch Evidence for Generalizable AI-Generated Image Detection}

\author{
Shengbo Qi\textsuperscript{\rm 1},
Hongyi Fang\textsuperscript{\rm 1},
Benjia Zhou\textsuperscript{\rm 1,}\thanks{Corresponding author.},
Rui Mao\textsuperscript{\rm 2}
}

\affiliations{
\textsuperscript{\rm 1}Beijing Institute of Technology, Zhuhai\\
\textsuperscript{\rm 2}Shenzhen University
}
\date{}
\begin{document}

\maketitle

\begin{abstract}

AI-generated image detectors generalize poorly when their training and test images originate from different generators or datasets. Despite the rich spatial representations produced by vision foundation models like  DINO, existing detectors typically classify images using only the globally aggregated CLS token. We hypothesize that globally aggregating DINO features into a single CLS token obscures spatially distributed generation traces. To test this hypothesis, we introduce PatchHead, a lightweight spatial aggregation head that preserves the two-dimensional organization of DINO patch tokens and integrates evidence across neighboring regions. 
During training, we freeze the pretrained DINO backbone and optimize only the inserted LoRA adapters, PatchHead, and auxiliary projection head.
Across nine cross-dataset benchmarks spanning manually curated and in-the-wild settings, PatchHead ranks first on seven datasets and second on the remaining two. It improves the strongest prior method from $91.6\%$ to $94.6\%$ in average balanced accuracy (\textbf{+3.0} points) and raises the worst-case accuracy from $82.4\%$
to $89.4\%$ (\textbf{+6.9} points), while introducing only \textbf{$8.6\%$} more trainable parameters and \textbf{$0.08\%$} additional FLOPs. Further qualitative analysis suggests that PatchHead (i) reduces class-conditional domain discrepancy, and (ii) redirects the representation from content-dominated saliency toward spatially distributed authenticity evidence. 
Together, these observations provide a representation-level account of why spatial patch aggregation transfers more reliably across generators and datasets than a single CLS-based global representation. Our code and models will be made available upon acceptance.

\end{abstract}

\section{Introduction}
AI-generated image detection (AIGID) is fundamentally an out-of-distribution recognition problem~\cite{DBLP:conf/nips/CaiWZYT25,DBLP:conf/cvpr/OjhaLL23,ren2026well}. A detector trained on a limited collection of generators must remain effective when confronted with unseen synthesis models, image sources, and post-processing pipelines.  Yet, despite near-saturated in-distribution performance, existing detectors often suffer substantial
degradation once the training and test distributions differ ~\cite{DBLP:conf/nips/ZhuCYHLLT0H023,DBLP:conf/iclr/YanLCHJ0X25,DBLP:journals/corr/abs-2406-09398}. This gap suggests that improving the backbone alone may be insufficient; how its visual representations are used for authenticity prediction may be equally important. 

A common design is to use only the final CLS token for downstream classification, leaving the spatial organization of patch tokens unexploited at the detector interface~\cite{DBLP:conf/iclr/DosovitskiyB0WZ21,dhakal2026simlbr,DBLP:conf/cvpr/GuillaroZUSCV25,DBLP:journals/corr/abs-2602-02222}. This choice is natural for semantic recognition, where a global representation summarizes what is present in an image. Its suitability for AIGID, however, is less obvious: the detector must identify how an image was produced rather than what it depicts. Generation-related evidence can be weak, localized, and distributed across multiple regions, and may not coincide with the most semantically salient objects~\cite{zhong2023patchcraft}. Compressing such evidence into a single global token may therefore attenuate authenticity-related cues while retaining image-content and dataset-specific variations. This raises a simple question: \textit{is the globally aggregated CLS token the appropriate representation for generalizable AI-generated image detection?}

\begin{figure}[t]
    \centering
    \includegraphics[
        width=0.9\linewidth
    ]{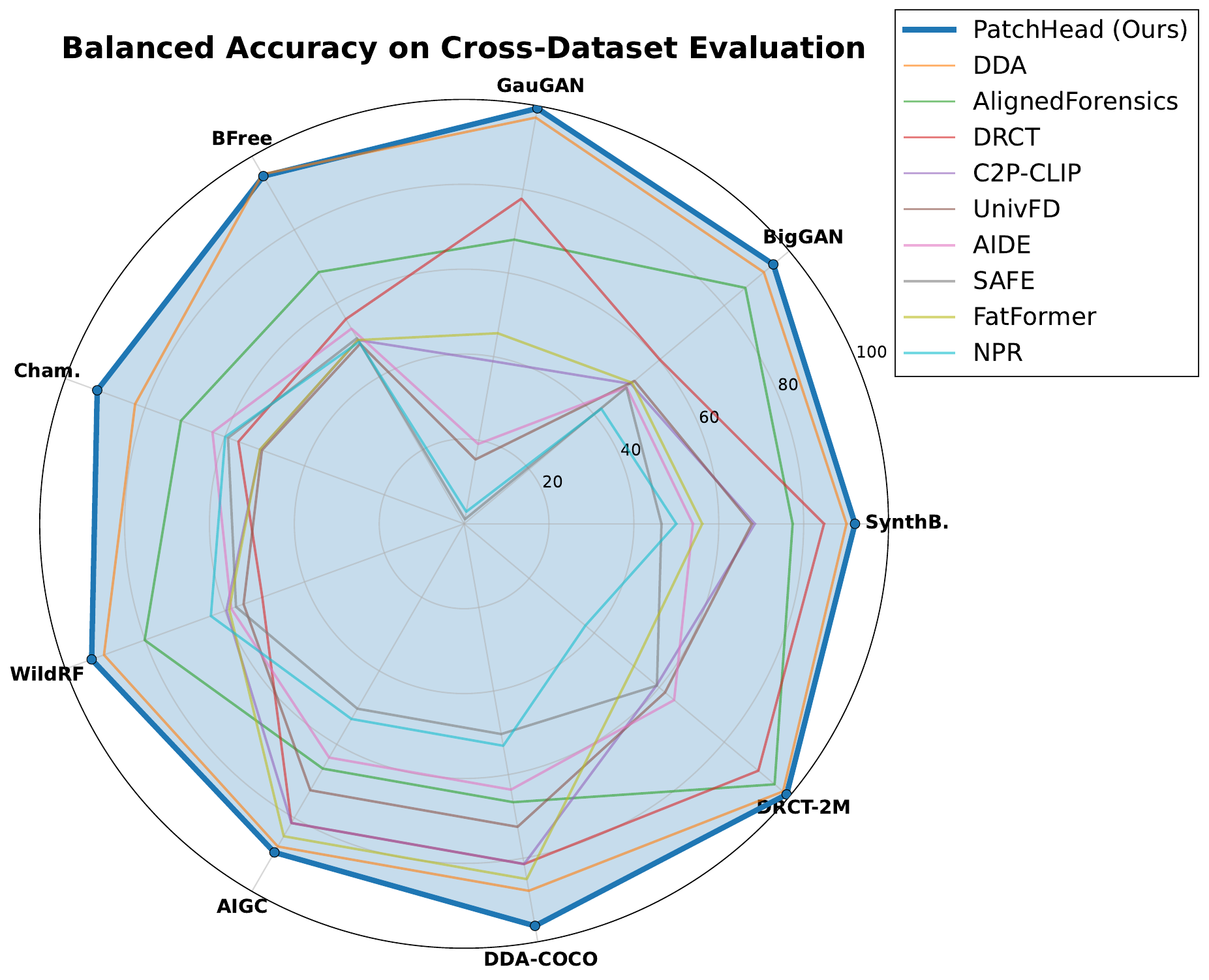}
    \caption{
    Radar visualization of balanced accuracy on cross-dataset evaluation.
    The proposed PatchHead consistently achieves the best performance
    across most generative categories.
    }
    \label{fig:cross_dataset_radar}
\end{figure}

We hypothesize that the limited cross-domain generalization of CLS-based detectors stems partly from a representation mismatch: global aggregation entangles authenticity prediction with content- and dataset-specific variations, while weakening spatially distributed generation evidence. 
To test this hypothesis, we introduce PatchHead, a lightweight replacement for the conventional CLS-based classifier. We keep the pretrained DINO backbone~\cite{ DBLP:journals/tmlr/OquabDMVSKFHMEA24,DBLP:journals/tmlr/SimeoniVSBOJKSYRMHWWDMS26} and the remaining training protocol unchanged. PatchHead reshapes the output patch tokens according to their original two-dimensional layout, models interactions among neighboring regions with a small convolutional module, and aggregates the resulting spatial evidence for image-level prediction. 
Our goal is not to design a new vision backbone or a sophisticated forensic architecture. Instead, we isolate a single design choice: \textit{whether DINO representations should be globally compressed into a CLS token or spatially modeled before image-level aggregation.}


Despite its minimal design, PatchHead delivers strong cross-dataset
generalization, as shown in Fig.~\ref{fig:cross_dataset_radar}. Under a
single-model evaluation protocol without target-specific fine-tuning or
test-time adaptation, it ranks first on seven of nine benchmarks and
second on the remaining two. Compared with the strongest prior method,
PatchHead improves the average balanced accuracy from $91.6\%$ to
$94.6\%$ (\textbf{+3.0} points) and raises the worst-case accuracy from
$82.4\%$ to $89.4\%$ (\textbf{+6.9} points), while introducing only
\textbf{$8.6\%$} more trainable parameters and \textbf{$0.08\%$} additional FLOPs.
A series of controlled studies supports the underlying representation hypothesis.  Simply averaging patch tokens does not reproduce the gains, and randomizing their spatial arrangement substantially degrades performance. A 3-hidden-layer MLP head also underperforms PatchHead, indicating that the improvement cannot be explained by additional capacity alone. Feature-space analyses further show reduced class-conditional discrepancy between training and unseen test domains.

Together, these results suggest that the standard CLS-based interface underutilizes the spatial representations already available in DINO. For generalizable AIGID, patch tokens should be spatially modeled before being compressed into an image-level prediction.

Our contributions are threefold:
\begin{itemize}
    \item We identify and empirically characterize a representation
    mismatch between CLS-based global features and the spatially
    distributed evidence required for generalizable AI-generated image
    detection.

    \item We introduce PatchHead, a lightweight spatial aggregation head that
    preserves the two-dimensional organization of DINO patch tokens and
    models neighboring regional evidence while keeping the pretrained
    backbone unchanged.

    \item Extensive experiments across nine cross-dataset benchmarks, together with controlled ablations, feature-space measurements, and
    spatial-response analyses, demonstrate the effectiveness of spatial patch aggregation for cross-generator and cross-dataset generalization.
\end{itemize}

\begin{figure}[t]
    \centering
    \includegraphics[width=\linewidth]{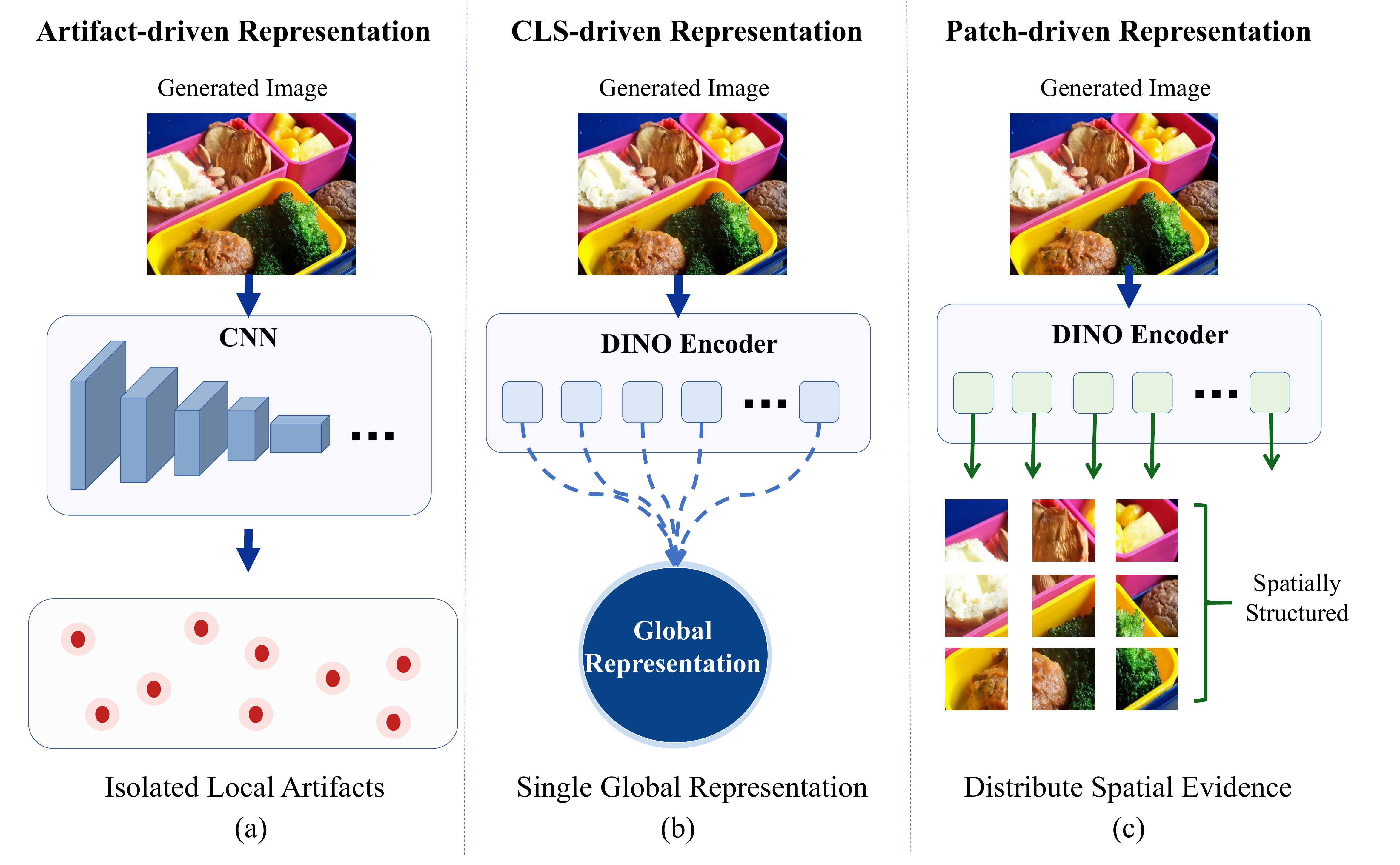}
    \caption{
    Conceptual comparison of representation strategies for
    AI-generated image detection.
    }
    \label{fig:model_comparison}
\end{figure}
\section{Related Work}

\subsection{Generalizable AI-Generated Image Detection}
\begin{figure*}[t]
    \centering
    \includegraphics[
        width=0.98\linewidth
    ]{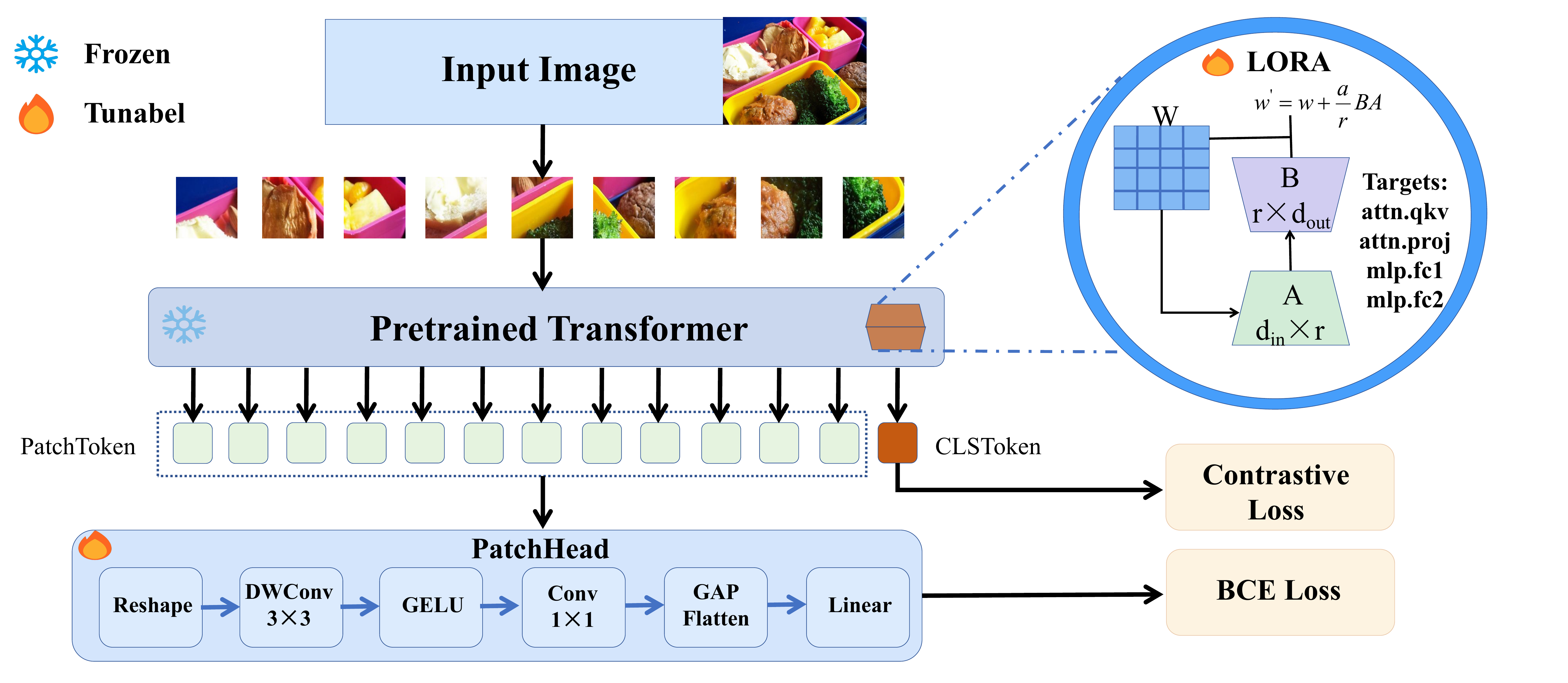}
    \caption{
    Overall framework of PatchHead.
    DINOv3 ViT-L/16 is adopted as the visual backbone and optimized with
    LoRA-based parameter-efficient fine-tuning.
    Unlike conventional CLS-based detectors, PatchHead utilizes Patch Tokens
    and preserves their two-dimensional spatial organization.
    The Patch Tokens are processed by a lightweight spatial aggregation head
    consisting of reshaping, depth-wise convolution, point-wise convolution,
    and global pooling operations.
    During training, CLS Token representations are only used for auxiliary
    contrastive learning and are discarded during inference.
    }
    \label{fig:patchhead}
\end{figure*}
Early general-purpose detectors learned generator fingerprints or frequency-sensitive artifacts from CNN-generated imagery \cite{DBLP:conf/cvpr/WangW0OE20}.
Subsequent approaches developed more explicit forms of generation evidence, including up-sampling-induced pixel dependencies~\cite{DBLP:conf/cvpr/TanLZWGLW24}, local texture statistics~\cite{zhong2023patchcraft}, diffusion reconstruction
residuals~\cite{DBLP:conf/iccv/WangBZWHCL23}, and challenging reconstruction-based training pairs~\cite{DBLP:conf/icml/ChenZYY24}. These methods improve generalization by explicitly designing or amplifying particular forensic cues, but their effectiveness still remains sensitive to generator architectures or image-processing conditions~\cite{ren2026well}.  

A complementary line of work focuses on reducing shortcut learning
induced by semantic bias and domain discrepancies in AI-generated image detection.B-Free constructs semantically aligned real--fake pairs~\cite{DBLP:conf/cvpr/GuillaroZUSCV25}, while Dual Data Alignment aligns real and generated images in both pixel and frequency domains ~\cite{DBLP:conf/nips/ChenXYZWXCXGYD25}.
These studies primarily improve which forensic evidence is learned or how biased training distributions are constructed. Our work instead investigates how pretrained spatial representations should be organized at the detector interface.

\begin{algorithm}[t]
\small
\caption{PyTorch-like pseudocode of PatchHead}
\label{alg:patchhead}
\begin{lstlisting}[style=pytorch]
class PatchHeadDetector(nn.Module):
    def __init__(self, dino, C, head_dim):
        super().__init__()
        self.backbone = AddLoRA(Freeze(dino))
        self.patch_head = nn.Sequential(
            DWConv3x3(groups=C),
            GELU(),
            Conv1x1(C, head_dim),
            AdaptiveAvgPool2d(1),
            Flatten(1)
            )
        self.classifier = Linear(head_dim, 1)
    def forward(I):
        z_cls, Z_p = self.backbone(I)
        B, N, C = Z_p.shape
        P = Reshape(Z_p, B, C, H, W)
        f_patch = self.patch_head(P)
        logit = self.classifier(f_patch)
        return logit, z_cls
# Optimization
L_cls = BCEWithLogits(logit, y)
L_con = Contrastive(z_cls)
L = (1-lambda) * L_cls + lambda * L_con
\end{lstlisting}
\end{algorithm}
\subsection{Vision Foundation Models for Detection}
Large-scale vision foundation models, including CLIP family~\cite{DBLP:conf/icml/RadfordKHRGASAM21, DBLP:conf/cvpr/OjhaLL23,DBLP:conf/cvpr/CozzolinoPCNV22} and the DINO family~\cite{ DBLP:journals/tmlr/OquabDMVSKFHMEA24,DBLP:journals/tmlr/SimeoniVSBOJKSYRMHWWDMS26} have become strong feature extractors for generalizable detection.
Recent detectors use these foundations in different ways: B-Free~\cite{DBLP:conf/cvpr/GuillaroZUSCV25} fine-tunes DINOv2 features, SimLBR~\cite{dhakal2026simlbr} regularizes DINOv3 latent geometry, and MIRROR~\cite{DBLP:journals/corr/abs-2602-02222} compares DINOv3 features against a learned real-image manifold.
Their success motivates our backbone choice, but does not establish that a single global token is the best detector input.

\subsection{Patch-Level and Spatial Representation Learning}

Patch-level processing has been explored through local crops, texture
selection, and input-space patch
rearrangement~\cite{zhong2023patchcraft,DBLP:conf/nips/ZhengL0WGL024}. 
These methods operate directly on image regions or treat local patches as candidate
forensic cues. Recent work has further incorporated patch tokens from
vision foundation models through learnable attention pooling, combining
global and local token representations for image-level prediction
~\cite{Abdullah_2026_CVPR}. 
\method{} explores a different perspective.
Rather than aggregating patch tokens primarily as a token sequence into a global embedding, it restores their explicit two-dimensional layout and models interactions according to spatial adjacency before image-level
aggregation.

\section{Method}
Existing detectors typically obtain authenticity evidence either from explicitly designed forensic cues (Fig.~\ref{fig:model_comparison}(a))~\cite{DBLP:conf/iccv/WangBZWHCL23,DBLP:conf/cvpr/0005LW0S25,DBLP:conf/cvpr/TanLZWGLW24} or from a single globally aggregated representation (Fig.~\ref{fig:model_comparison}(b))~\cite{DBLP:conf/cvpr/GuillaroZUSCV25,dhakal2026simlbr,DBLP:journals/corr/abs-2602-02222}.
PatchHead (Fig.~\ref{fig:model_comparison}(c)) instead preserves the native spatial layout of foundation-model patch tokens and models neighboring evidence before image-level aggregation. We first formulate the cross-domain detection problem and
then introduce the proposed spatial prediction head.
\paragraph{Problem Formulation}
Let $\mathcal{D}_{s}=\{(x_i,y_i)\}_{i=1}^{N}$ denote a source-domain
training set, where $x_i$ is an image and
$y_i\in\{0,1\}$ indicates whether the image is real or synthetic.
The detector is evaluated on unseen target domains
$\mathcal{D}_{t}$ whose image sources, generator families, or
post-processing pipelines may differ from those observed during training.

Given an input image $x$, a pretrained DINO backbone produces one global
CLS token and a set of patch tokens:
\begin{equation}
    \mathbf{Z}(x)
    =
    [\mathbf{z}_{\mathrm{cls}},
    \mathbf{z}_{1},\ldots,\mathbf{z}_{N_p}],
\end{equation}
where $\mathbf{z}_{\mathrm{cls}}\in\mathbb{R}^{D}$ and
$\mathbf{z}_{i}\in\mathbb{R}^{D}$.
Our objective is to learn an image-level authenticity predictor that
generalizes from $\mathcal{D}_{s}$ to unseen target domains without
target-domain adaptation.

\begin{table*}[t]
\centering

\fontsize{7}{8}\selectfont
\setlength{\tabcolsep}{1.5pt}

\resizebox{\textwidth}{!}{
\begin{tabular}{lccccccccc>{\columncolor{gray!15}}cc}
\toprule

\multirow{2}{*}{Method} &
\multicolumn{5}{c}{\textbf{Manually Curated Datasets}} &
\multicolumn{4}{c}{\textbf{In-the-Wild Datasets}} &
\multirow{2}{*}{Avg.} &
\multirow{2}{*}{Min.} 
\\

\cmidrule(lr){2-6}
\cmidrule(lr){7-10}

&DRCT-2M 
&DDA-COCO 
&EvalGEN 
&SynthB. 
&AIGC 
&Cham. 
&SynthW. 
&WildRF 
&BFree &
&
\\

\midrule

NPR~\cite{DBLP:conf/cvpr/TanLZWGLW24}
&37.3&42.2&2.9&50.0&53.1&59.9&49.8&63.5&49.5&45.4&2.9\\

UnivFD~\cite{DBLP:conf/cvpr/OjhaLL23}
&61.8&52.4&15.4&67.8&72.5&50.7&52.3&55.3&49.0&53.0&15.4
\\

FatFormer~\cite{DBLP:conf/cvpr/LiuTTW0Z24}
&52.2&51.7&45.6&56.1&85.0&51.2&52.1&58.9&50.0&55.9&45.6
\\

SAFE~\cite{DBLP:conf/kdd/LiCHJHF25}
&59.3&49.9&1.1&46.5&50.3&59.2&49.1&54.2\rlap{$^*$}&50.5&46.7&1.1
\\

C2P-CLIP~\cite{DBLP:conf/aaai/TanTLGWZW25}
&59.2&51.3&38.9&68.5&81.4&51.1&57.1&59.6&50.0&57.5&38.9
\\

AIDE~\cite{DBLP:conf/iclr/YanLCHJ0X25}
&64.6&50.0&19.1&53.9&63.6&63.1&54.7\rlap{$^*$}&58.4&53.1&53.4&19.1
\\

DRCT~\cite{DBLP:conf/icml/ChenZYY24}
&90.5&60.2&77.8&81.3\rlap{$^*$}&81.4&56.6&55.1&51.6\rlap{$^*$}&55.7&67.8&51.6
\\

AlignedForensics~\cite{DBLP:conf/iclr/RajanOSL25}
&95.5&86.5&68.0&77.4&66.6&71.0&78.8&80.1&68.5&76.9&66.6
\\

DDA~\cite{DBLP:conf/nips/ChenXYZWXCXGYD25}
& \underline{98.1} & \underline{92.2} & \underline{97.2} & \underline{90.1} & \underline{87.8} & \underline{82.4} & \textbf{90.9}&\underline{90.3}&\textbf{95.1}&\underline{91.6}&\underline{82.4}
\\

\midrule


PatchHead(ours)
&\textbf{99.2}
&\textbf{95.1}
&\textbf{99.4}
&\textbf{92.1}
&\textbf{96.2}
&\textbf{92.0}
&\underline{89.4}
&\textbf{93.4}
&\underline{94.6}
&\textbf{94.6}
&\textbf{89.4}
\\

\bottomrule

\end{tabular}
}
\caption{
Balanced Accuracy (\%) on cross-dataset evaluation.
The best and second-best results for each generator are highlighted in
\textbf{bold} and \underline{underline}, respectively. *  indicates values corrected from the original reports (DDA~\cite{DBLP:conf/nips/ChenXYZWXCXGYD25}). Avg. and Min. denote the average and minimum accuracy across all datasets, respectively.
}
\label{tab:cross_dataset}
\end{table*}

\paragraph{PatchHead}
A conventional DINO-based detector predicts image authenticity from the
global CLS token:
\begin{equation}
    \hat y_{\mathrm{cls}}
    =
    g_{\mathrm{cls}}(\mathbf z_{\mathrm{cls}}),
\end{equation}
where $ g_{\mathrm{cls}}(\cdot)$ represents a nonlinear mapping function.
PatchHead instead retains patch tokens as spatially indexed regional
representations. (Fig.~\ref{fig:patchhead})

Let $N_p=H_pW_p$. We restore the token sequence to its
original spatial layout:
\begin{equation}
    \mathbf F_0
    =
    \operatorname{Reshape}(\mathbf Z_p)
    \in
    \mathbb R^{D\times H_p\times W_p}.
\end{equation}
PatchHead first applies a  $3\times3$ depthwise convolution~\cite{DBLP:journals/corr/HowardZCKWWAA17} to exchange
information among spatially adjacent patches, followed by a nonlinear
activation and a $1\times1$ pointwise projection:
\begin{equation}
    \mathbf F_h
    =
    \operatorname{Conv}_{1\times1}
    \left(
    \operatorname{GELU}
    \left(
    \operatorname{DWConv}_{3\times3}(\mathbf F_0)
    \right)
    \right).
\end{equation}
The depthwise convolution models local spatial interactions, whereas the
pointwise convolution mixes feature channels and projects the DINO
dimension to the detector-head dimension.
The image-level patch representation is obtained through global average
pooling:
\begin{equation}
    \mathbf f_{\mathrm{patch}}
    =
    \operatorname{GAP}(\mathbf F_h),
\end{equation}
and the authenticity probability is predicted as
\begin{equation}
    \hat y
    =
    \sigma
    \left(
    \mathbf w^\top\mathbf f_{\mathrm{patch}}+b
    \right).
\end{equation}

\paragraph{Training Objective}
\method{} is trained with image-level classification supervision and an
auxiliary contrastive objective as shown in Algorithm~\ref{alg:patchhead}. The final prediction relies exclusively
on the spatial patch representation:
\begin{equation}
    \mathcal{L}_{\mathrm{cls}}
    =
    -y\log\hat{y}
    -(1-y)\log(1-\hat{y}),
\end{equation}
where $y$ is the ground-truth label and $\hat{y}$ is the predicted
probability of being synthetic.

Following prior work~\cite{DBLP:conf/nips/ChenXYZWXCXGYD25}, we freeze the original DINO
weights and optimize its LoRA adapters. An auxiliary contrastive loss
$\mathcal{L}_{\mathrm{con}}$ is applied to the CLS representations of
augmented views to regularize the adapted backbone. The CLS branch is
used only during training and discarded at inference.

The overall objective is
\begin{equation}
    \mathcal{L}
    =
    (1-\lambda)\mathcal{L}_{\mathrm{cls}}
    +
    \lambda\mathcal{L}_{\mathrm{con}},
\end{equation}
where $\lambda$ balances the two objectives.

\section{Experiments}




















\paragraph{Datasets.}
We evaluate PatchHead on nine benchmarks, including five curated datasets
(DRCT-2M~\cite{DBLP:conf/icml/ChenZYY24}, DDA-COCO~\cite{DBLP:conf/nips/ChenXYZWXCXGYD25}, EvalGEN~\cite{DBLP:conf/nips/ChenXYZWXCXGYD25}, Synthbuster~\cite{synthbuster}, and
AIGCDetectionBenchmark~\cite{zhong2023patchcraft}) and four in-the-wild datasets
(Chameleon~\cite{DBLP:conf/iclr/YanLCHJ0X25}, SynthWildX~\cite{DBLP:conf/cvpr/CozzolinoPCNV22}, WildRF~\cite{DBLP:journals/corr/abs-2406-09398}, and BFree-Online~\cite{DBLP:conf/cvpr/GuillaroZUSCV25}). The curated benchmarks
cover diverse generative paradigms and provide controlled evaluation of
cross-generator generalization, whereas the in-the-wild benchmarks
contain online images with more diverse content, post-processing, and
generation sources. Detailed statistics are provided in the supplementary material.



\begin{table*}[t]
\centering

\scriptsize
\fontsize{7}{8}\selectfont
\setlength{\tabcolsep}{1.5pt}

\resizebox{\textwidth}{!}{
\begin{tabular}{lccccccccccccccccc>{\columncolor{gray!15}}c}
\toprule

Method
& ADM
& D2
& Glide
& MJ
& VQDM
& BigGAN
& CycleGAN
& GauGAN
& ProGAN
& SDXL
& SD1.4
& SD1.5
& StarGAN
& StyleGAN
& StyleGAN2
& WFR
& Wukong
& Avg.
\\

\midrule

NPR~\cite{DBLP:conf/cvpr/TanLZWGLW24}
&43.8&20.0&41.2&53.4&48.4&53.1&76.6&42.2&58.7&59.6&55.1&55.0&67.4&57.9&54.6&58.8&57.4&53.1\\

UnivFD~\cite{DBLP:conf/cvpr/OjhaLL23}
&62.5&50.0&61.3&55.1&76.9&87.5&\underline{96.9}&\underline{98.8}&\textbf{99.4}&58.2&55.6&55.7&95.1&80.0&69.4&69.2&61.1&72.5\\

FatFormer~\cite{DBLP:conf/cvpr/LiuTTW0Z24}
&80.2&68.5&\underline{91.1}&54.4&88.0&\textbf{99.2}&\textbf{99.5}&\textbf{99.1}&98.5&71.7&67.5&67.2&\underline{99.4}&\textbf{98.0}&\textbf{98.8}&88.3&75.6&85.0\\

SAFE~\cite{DBLP:conf/kdd/LiCHJHF25}
&49.5&49.5&53.0&49.0&50.2&52.2&51.9&50.0&50.0&49.8&49.7&49.8&50.1&50.0&50.0&49.8&50.3&50.3\\

C2P-CLIP~\cite{DBLP:conf/aaai/TanTLGWZW25}
&71.6&52.3&73.5&56.6&73.7& \underline{98.4}&96.8&\underline{98.8}&\underline{99.3}&62.3&77.5&76.9&\textbf{99.6}&\underline{93.1}&79.4&\underline{94.8}&79.4&81.4\\

AIDE~\cite{DBLP:conf/iclr/YanLCHJ0X25}
&52.9&51.1&60.2&49.8&69.3&70.1&93.6&60.6&89.0&49.6&51.6&51.0&72.1&66.5&59.0&80.6&54.5&63.6\\

DRCT~\cite{DBLP:conf/icml/ChenZYY24}
&79.9&89.2&89.2&85.5&\underline{88.6}&81.4&91.0&93.8&71.1&88.3&91.4&91.0&53.0&62.7&63.8&73.9&90.8&81.4\\

AlignedForensics~\cite{DBLP:conf/iclr/RajanOSL25}
&51.6&52.0&55.6&\underline{96.2}&72.1&51.2&49.5&50.8&50.7&95.1&\textbf{99.7}&\textbf{99.6}&53.8&52.7&51.6&50.0&\textbf{99.6}&66.6\\

DDA~\cite{DBLP:conf/nips/ChenXYZWXCXGYD25}
&\underline{89.5}&\underline{94.6}&89.6&95.6&76.6&91.0&72.5&92.7&92.8&\underline{99.4}&98.7&98.6&72.7&87.8&90.2&52.1&98.8&\underline{87.8} \\

\midrule


PatchHead(ours)
&\textbf{95.5}
&\textbf{99.2}
&\textbf{96.7}
&\textbf{97.9}
&\textbf{98.0}
&97.1
&92.6
&97.8
&94.2
&\textbf{99.5}
&\underline{98.9}
&\underline{99.0}
&84.9
&92.7
&\underline{94.1}
&\textbf{98.7}
&\underline{99.1}
&\textbf{96.2}
\\

\bottomrule
\end{tabular}
}
\caption{
Balanced Accuracy (\%) on the AIGCDetectionBenchmark.
}
\label{tab:aigc}
\end{table*}

\begin{table*}[t]
\centering

\small
\setlength{\tabcolsep}{3pt}

\begin{tabular*}{\textwidth}{@{\extracolsep{\fill}}lccc>{\columncolor{gray!15}}cccc>{\columncolor{gray!15}}c@{}}
\toprule

\multirow{2}{*}{Method}
&
\multicolumn{4}{c}{\textbf{SynthWildX}}
&
\multicolumn{4}{c}{\textbf{WildRF}}
\\

\cmidrule(lr){2-5}
\cmidrule(lr){6-9}

&DALL-E 3
&Firefly
&Midjourney
&Avg.
&Facebook
&Reddit
&Twitter
&Avg.
\\

\midrule

NPR~\cite{DBLP:conf/cvpr/TanLZWGLW24}
&43.6&61.3&44.5&49.8&78.1&61.0&51.3&63.5\\

UnivFD~\cite{DBLP:conf/cvpr/OjhaLL23}
&45.4&65.3&46.2&52.3&49.1&60.2&56.5&55.3\\

FatFormer~\cite{DBLP:conf/cvpr/LiuTTW0Z24}
&46.5&61.6&48.3&52.1&54.1&68.1&54.4&58.9\\

SAFE~\cite{DBLP:conf/kdd/LiCHJHF25}
&49.4&48.2&49.6&49.1&50.9&74.1&37.5&54.2\rlap{$^*$}\\

C2P-CLIP~\cite{DBLP:conf/aaai/TanTLGWZW25}
&56.9&61.4&53.0&57.1&54.4&68.4&55.9&59.6\\

AIDE~\cite{DBLP:conf/iclr/YanLCHJ0X25}
&63.4&48.8&51.9&54.7\rlap{$^*$}&57.8&71.5&45.8&58.4\\

DRCT~\cite{DBLP:conf/icml/ChenZYY24}
&58.3&56.4&50.5&55.1&46.6&53.1&55.2&51.6\rlap{$^*$}\\

AlignedForensics~\cite{DBLP:conf/iclr/RajanOSL25}
&85.5&58.5&\underline{92.2}&78.8&89.4&69.1&81.8&80.1\\

DDA~\cite{DBLP:conf/nips/ChenXYZWXCXGYD25}
&\textbf{92.3}&\underline{87.3}&\textbf{93.1}&\textbf{90.9}&\underline{93.1}&\underline{86.4}&\underline{91.5}&\underline{90.3}\\

\midrule


PatchHead(ours)
&\underline{89.8}
&\textbf{88.3}
&90.0
&\underline{89.4}
&\textbf{93.4}
&\textbf{94.0}
&\textbf{92.8}
&\textbf{93.4}
\\

\bottomrule
\end{tabular*}
\caption{
Balanced Accuracy (\%) on SynthWildX and WildRF.
}
\label{tab:synthwildx_wildrf}
\end{table*}

\paragraph{Training and Evaluation Protocol.} 
We use DINOv3 ViT-L/16~\cite{ DBLP:journals/tmlr/SimeoniVSBOJKSYRMHWWDMS26} with frozen base parameters and trainable LoRA
adapters of rank 8. All models are trained for one epoch on the
DDA-aligned COCO-SD-2 dataset~\cite{DBLP:conf/nips/ChenXYZWXCXGYD25} using the same augmentations, optimization
schedule, and auxiliary contrastive objective. Training augmentations
include random cropping, pixel- and frequency-domain mixing, random JPEG
compression, and multi-scale augmentations via random downsampling and upsampling for contrastive learning.
Center cropping is used at inference. Unless otherwise specified, the
contrastive-loss weight is set to $\lambda=0.5$ , as determined by ablation experiments. We evaluate a single trained model on all target benchmarks without dataset-specific fine-tuning or test-time adaptation, and report balanced accuracy as the primary metric. Full implementation details are provided in the
supplementary material.


\begin{table*}[t]
\centering

\small
\setlength{\tabcolsep}{3pt}

\begin{tabular}{lccccccccc>{\columncolor{gray!15}}c c c}
\toprule

Representation
&Cham.
&AIGC
&DRCT-2M
&SynthB.
&SynthW.
&WildRF
&BFree
&DDA-COCO
&EvalGEN
&Avg.
&Params.
&FLOPs
\\

\midrule

CLSHead(Linear)
&91.3
&94.7
&99.0
&91.8
&88.9
&91.5
&92.5
&95.8
&\textbf{99.5}
&93.9
&3.15M
&$\sim$2K
\\

CLSHead(MLP)
&91.0
&95.2
&99.1
&90.6
&88.4
&93.3
&92.6
&95.4
&99.3
&93.9
&3.84M
&$\sim$1.38M
\\
GAP+Linear
&90.8
&94.8
&99.0
&91.4
&87.5
&92.3
&92.1
&95.7
&99.2
&93.6
&3.15M
&$\sim$0.45M
\\

PatchHead+Shuffle
&92.0
&95.7
&99.1
&91.2
&89.3
&\textbf{93.4}
&92.8
&\textbf{96.1}
&99.3
&94.3
&3.42M
&$\sim$242M
\\

\method{}
&\textbf{92.0}
&\textbf{96.2}
&\textbf{99.2}
&\textbf{92.1}
&\textbf{89.4}
&\textbf{93.4}
&\textbf{94.6}
&95.1
&99.4
&\textbf{94.6}
&3.42M
&$\sim$242M
\\

\bottomrule

\end{tabular}
\caption{
Ablation of representation interfaces and spatial organization.
All variants share the same DINOv3 backbone, LoRA configuration, training
data, augmentations, and contrastive objective.
PatchHead+Shuffle applies a fixed random permutation to patch positions
before spatial aggregation.The best results are shown in \textbf{bold}.
Params. denotes the total trainable parameters, and FLOPs denotes the
additional computational cost of the prediction head.
}
\label{tab:patchhead_analysis}
\end{table*}

\paragraph{Comparison with State-of-the-Art Methods}
As shown in Table~\ref{tab:cross_dataset}, \method{} ranks first on seven
of the nine unseen benchmarks and second on the remaining two. It achieves
an average balanced accuracy of $94.6\%$, outperforming the strongest
baseline, DDA, by $3.0$ percentage points, with consistent gains on both
manually curated ($+3.3$) and in-the-wild ($+2.7$) datasets. Moreover,
\method{} raises the worst-case accuracy from $82.4\%$ to $89.4\%$
(\textbf{$+6.9$} points), demonstrating more reliable generalization
across heterogeneous target domains and supporting spatial patch
aggregation as an effective detection interface.

\begin{figure*}[t]
    \centering
    
    \begin{subfigure}[t]{0.58\textwidth}
        \centering
        \includegraphics[width=\linewidth]{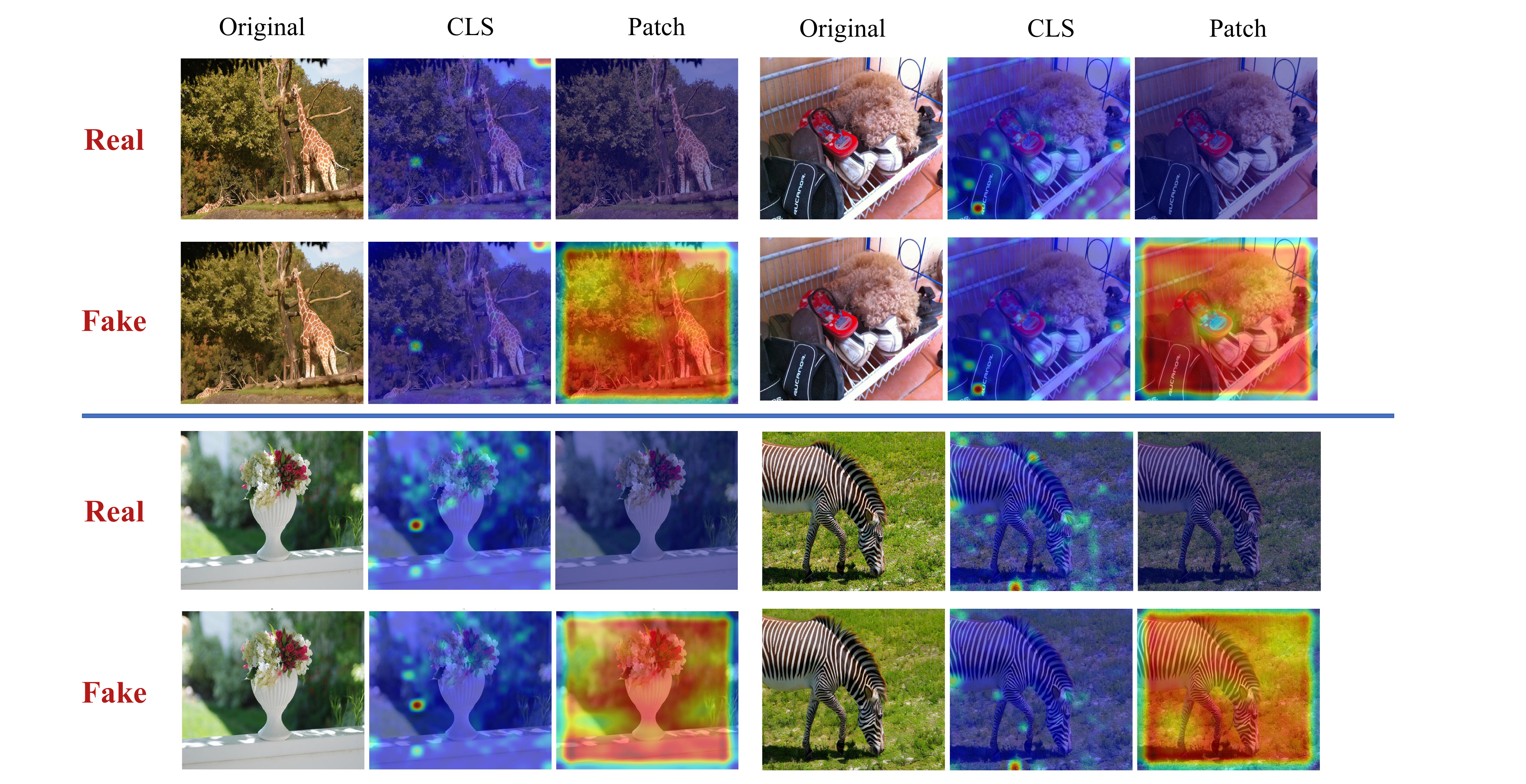 }
        \caption{
        Spatial response maps.
        }
        \label{fig:heatmap}
    \end{subfigure}
    \hfill
    \begin{subfigure}[t]{0.41\textwidth}
        \centering
        \includegraphics[width=\linewidth]{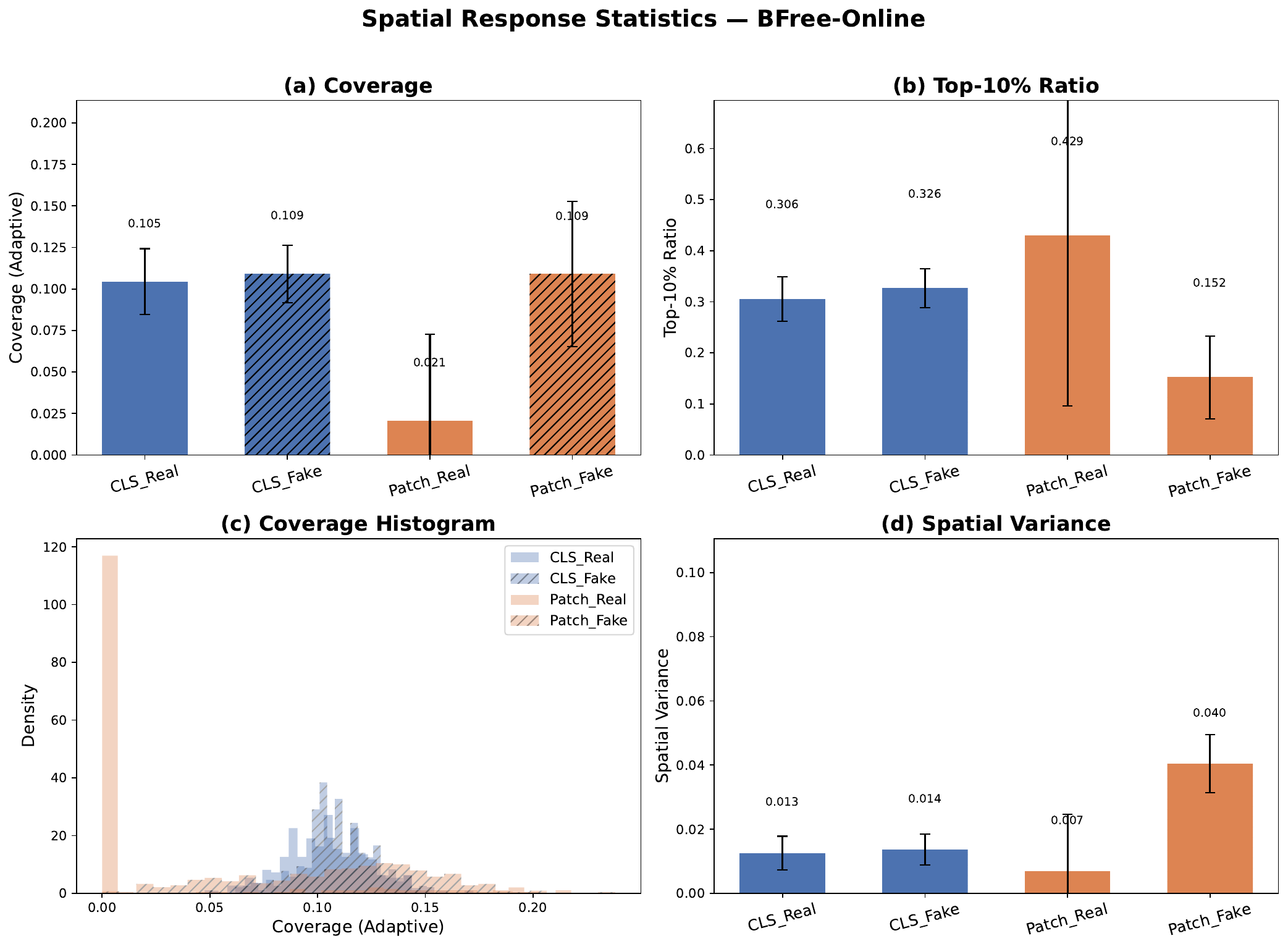}
        \caption{
        Response statistics on BFree-Online.
        }
        \label{fig:spatial_analysis}
    \end{subfigure}
    
    \caption{
    Qualitative and quantitative comparisons of spatial responses.
    Left: representative response maps of the CLS-based baseline and
    PatchHead on real and generated images.
    Right: dataset-level statistical analysis on BFree-Online, confirming
    that PatchHead produces more authenticity-sensitive and spatially
    distributed response patterns.
    }
    \label{fig:spatial_response_combined}
\end{figure*}

\begin{table*}[!ht]
\centering

\small
\setlength{\tabcolsep}{3.6pt}

\begin{tabular}{lccccccccc>{\columncolor{gray!15}}cc}
\toprule
Patch Aggregation Head
& Cham.
& AIGC
& DRCT-2M
& SynthB.
& SynthW.
& WildRF
& BFree
& DDA-COCO
& EvalGEN
& Avg.
& Min.
\\
\midrule


\method{}
&\textbf{92.0}
&\textbf{96.2}
&\textbf{99.2}
&92.1
&89.4
&93.4
&\textbf{94.6}
&95.1
&\textbf{99.4}
&\textbf{94.6}
&\textbf{89.4}
\\

\quad w/ Multi-scale
&91.5
&95.0
&99.0
&90.9
&88.1
&92.7
&93.7
&\textbf{95.8}
&99.3
&94.0
&88.1
\\

\quad w/  Multi-level
&85.4
&93.2
&99.0
&\textbf{95.3}
&\textbf{92.4}
&\textbf{94.0}
&93.4
&95.1
&97.0
&93.9
&85.4
\\

\quad w/ HRM~\cite{wang2025hierarchical}
&81.5
&89.2
&97.6
&87.2
&78.6
&84.4
&74.5
&90.9
&97.4
&86.8
&74.5
\\

\bottomrule
\end{tabular}
\caption{
Comparison of different patch aggregation heads.
All variants share the same DINOv3 backbone, LoRA configuration,
training objective, and classifier, and differ only in how spatial patch
features are aggregated.
}
\label{tab:sprm_ablation}
\end{table*}

\paragraph{Fine-grained comparison.}
Tables~\ref{tab:aigc} and~\ref{tab:synthwildx_wildrf} provide
generator- and source-level results on representative curated and
in-the-wild benchmarks. Detailed results on the remaining benchmarks are provided in the supplementary material. 
On AIGCDetectionBenchmark, \method{} achieves the best average accuracy of $96.2\%$, with notable gains of $9.4$ points on VQDM and $3.9$ points on WFR. On the in-the-wild benchmarks, it reaches $89.4\%$
on SynthWildX and improves DDA from $90.3\%$ to $93.4\%$ on WildRF,
ranking first across all three social-media sources. These results show
that the overall gains extend across diverse generators and online data
sources rather than being driven by a single subset.

\begin{figure*}[t]
    \centering
    \includegraphics[
        width=0.9\linewidth
    ]{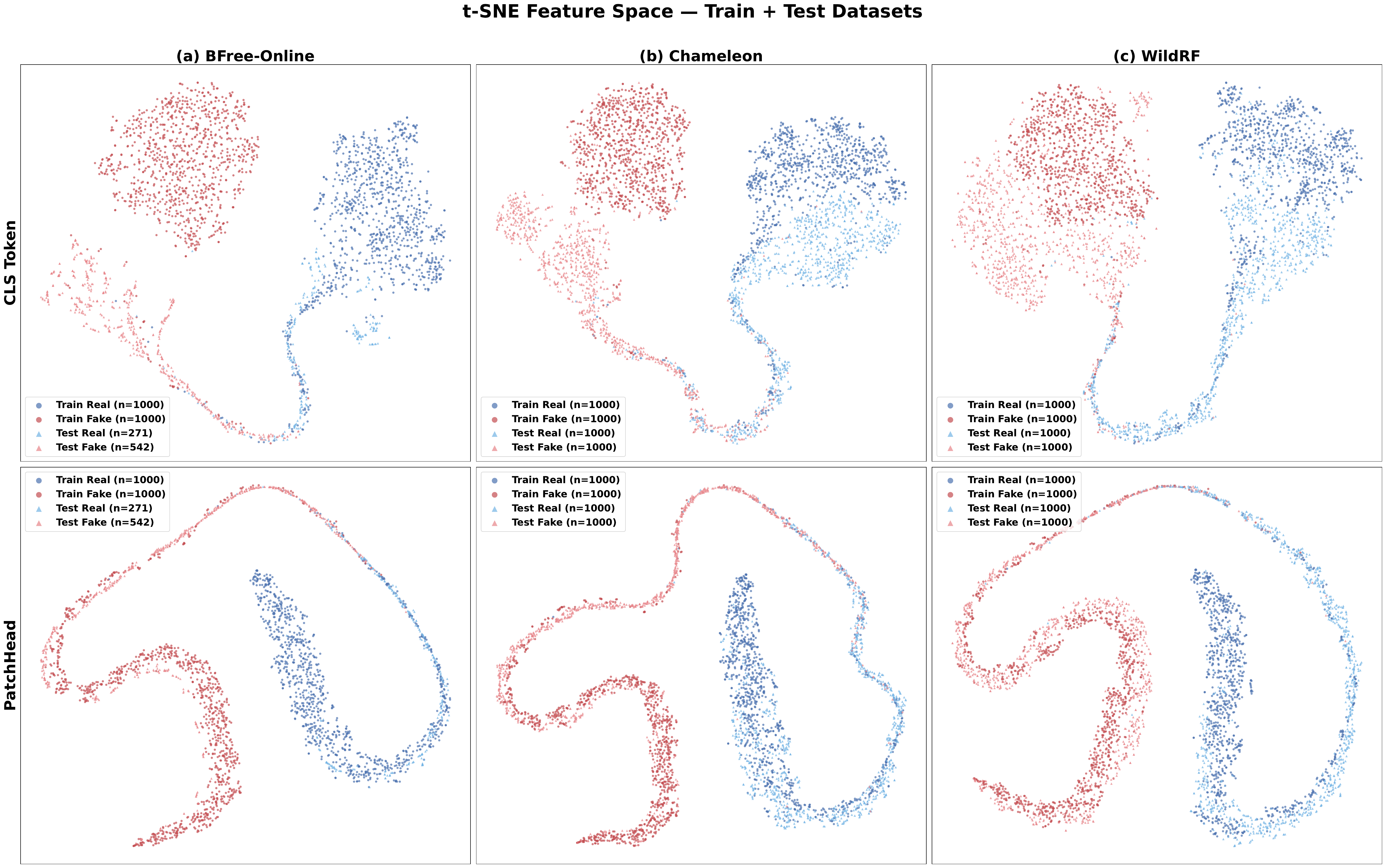}
    \caption{
    t-SNE visualization of feature distributions learned by the CLS-based baseline and PatchHead on the training set and the unseen
    test sets. CLS-based features remain strongly separated by dataset origin, whereas PatchHead exhibits greater class-conditional alignment across domains while preserving real--fake discrimination.
    }
    \label{fig:tsne_scatter_all}
\end{figure*}
\subsection{Ablation Studies}
\paragraph{\method{}  $vs.$ Alternative Representation Interfaces.} 

Table~\ref{tab:patchhead_analysis} shows that simple global averaging of
patch tokens performs slightly worse than the CLS baseline, indicating
that merely replacing CLS with patch features does not automatically
improve generalization. In contrast, \method{} achieves $94.6\%$ average
accuracy, outperforming CLSHead and GAP+Linear by $0.7$ and $1.0$ points,
respectively. Shuffling patch positions further reduces the result to
$94.3\%$. Thus, the main gain comes from learnable regional aggregation,
while the native spatial arrangement provides an additional benefit.

\paragraph{Authenticity-Adaptive Spatial Responses.} 
Figure~\ref{fig:heatmap} shows that the CLS-based detector responds to
similar sparse and content-salient regions in both real and generated
images, whereas \method{} exhibits more restricted responses on real
images and broader activations on generated images. Dataset-level
statistics on BFree-Online further support this observation
(Fig.~\ref{fig:spatial_analysis}): compared with CLS, \method{} produces
a larger real--synthetic coverage gap and a lower top-$10\%$ activation
concentration on generated images. The coverage distributions and spatial
variance show that this distinction persists across samples and response
structures. Together, these results suggest that \method{} aggregates
more distributed and authenticity-sensitive spatial evidence. Metric
definitions and additional results are provided in the supplementary
material.

\paragraph{Cross-Domain Feature Consistency.}
Figure~\ref{fig:tsne_scatter_all} visualizes the feature distributions of the
CLS-based baseline and PatchHead across the source training set and unseen
target datasets. The CLS-based representations form visibly separated
clusters according to dataset origin, even among samples sharing the same
authenticity label. In contrast, PatchHead brings real or synthetic
samples from different domains closer together while largely preserving
the separation between the two classes. This consistent pattern suggests
that spatial patch aggregation reduces dataset dependence and produces
more transferable representations for cross-domain authenticity
detection.

\begin{figure}[htbp]
    \centering
    \includegraphics[
        width=\linewidth
    ]{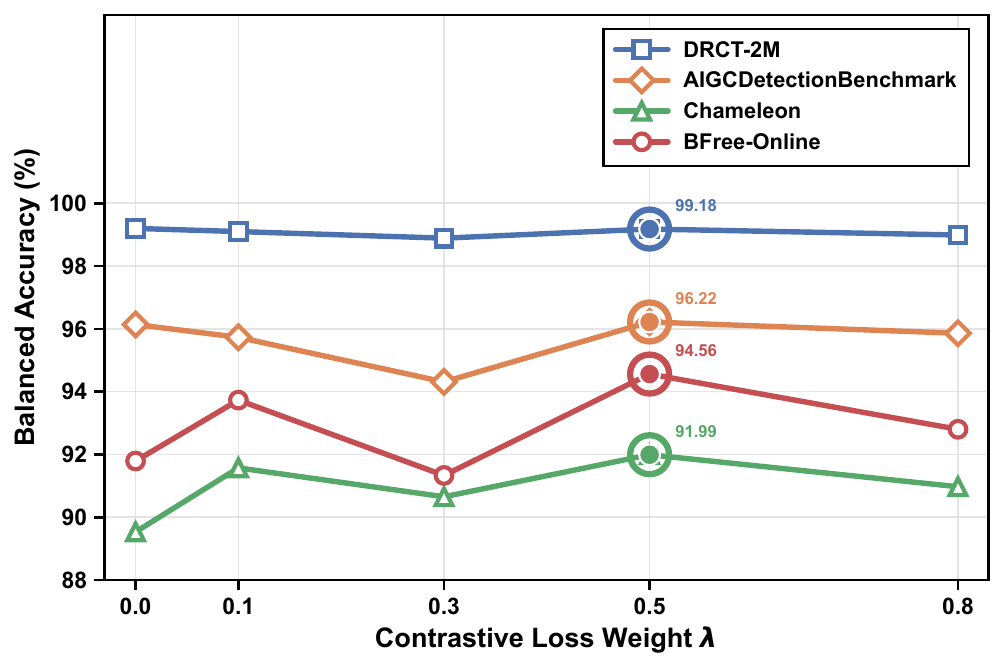}
    \caption{
    Effect of different contrastive learning weights for \method.
    }
    \label{fig:contrastive_weight}
\end{figure}
\paragraph{Is More Complex Spatial Designing Necessary?}
Table~\ref{tab:sprm_ablation} compares the default \method{} with
Multi-scale (uses multiple receptive fields), multi-level (fuses patch features from backbone layers 8, 12, 16, and 23), and HRM-inspired variants (introduces hierarchical aggregation) under identical settings. 
Detailed architectures are provided in the supplementary material.
The lightweight \method{} achieves the best
average and worst-case accuracies of $94.6\%$ and $89.4\%$, respectively.
More complex designs offer isolated gains but do not generalize
consistently, while the HRM-inspired variant drops to $86.8\%$ on average.
Thus, lightweight local aggregation provides the most reliable
cross-domain performance.

\paragraph{Sensitivity to Contrastive Regularization.} 

Figure~\ref{fig:contrastive_weight} evaluates the contrastive-loss weight
$\lambda$ on four representative curated and in-the-wild benchmarks, with
all other settings fixed. \method{} remains competitive across a broad
range of values, while $\lambda=0.5$ provides the best overall
performance. Larger weights yield no consistent gains, likely because
they reduce the relative contribution of authenticity classification.
We therefore adopt $\lambda=0.5$ throughout the experiments.

\begin{table}[htbp]
\centering

\small
\setlength{\tabcolsep}{5pt}
\begin{tabular}{lcccc}
\toprule
Method 
& Original
& JPEG60
& Resize2.0
& Blur2.0
\\
\midrule

CLSHead(Linear)
&93.9
&90.7
&93.6
&84.5
\\

PatchHead
&\textbf{94.6}
&\textbf{91.4}
&\textbf{94.6}
&\textbf{85.0}
\\

\bottomrule
\end{tabular}
\caption{
Robustness evaluation under different test-time perturbations.
Balanced accuracy (\%) is averaged over nine target benchmarks.
}
\label{tab:robustness}
\end{table}

\paragraph{Test-Time Robustness.}
As shown in Table~\ref{tab:robustness}, \method{} consistently
outperforms CLSHead under JPEG compression, resizing, and Gaussian blur.
It is particularly stable under resizing, retaining its original
$94.6\%$ accuracy. While blur remains challenging for both methods,
\method{} maintains higher performance under all tested perturbations.

\section{Conclusion}
This work revisits the representation interface of vision foundation models for generalizable AI-generated image detection. Instead of predicting authenticity from a single global CLS token, PatchHead restores DINO patch tokens to their native two-dimensional layout and applies lightweight local aggregation before image-level prediction. Controlled comparisons show that simply accessing patch tokens is insufficient: effective generalization requires learnable aggregation of regional representations, with their native spatial organization providing additional benefits. 
Feature-space and spatial-response analyses further suggest that PatchHead reduces dataset-specific biases while forming more distributed and authenticity-sensitive decision patterns.
These findings reveal that cross-domain transferability in foundation-model-based detectors depends not only on the choice of pretrained features, but also on preserving and aggregating their spatial structure.
Therefore, a minimal spatial inductive bias
can provide a stronger and more reliable detection interface than conventional global-token prediction.

\bibliography{references}
\clearpage
\appendix
\setcounter{secnumdepth}{1}
\begin{table*}[t!]
\centering
\small
\setlength{\tabcolsep}{5pt}

\begin{tabular}{llrrrrcl}
\toprule

Category &
Dataset &
Real &
Fake &
Images &
Generators &
Generator Families
\\

\midrule

\multirow{5}{*}{\textbf{Manually Curated}}
& DRCT-2M~\cite{DBLP:conf/icml/ChenZYY24}& 5,000& 80,000& 85,000& 16& Diffusion
\\

& DDA-COCO~\cite{DBLP:conf/nips/ChenXYZWXCXGYD25}& 5,000& 29,969& 34,969& 6& Diffusion
\\

& EvalGEN~\cite{DBLP:conf/nips/ChenXYZWXCXGYD25}& 0& 55,298& 55,298& 5& Flow Matching + Autoregressive
\\

& Synthbuster~\cite{synthbuster}& 1,000& 9,000& 10,000& 9& Diffusion
\\

& AIGC~\cite{zhong2023patchcraft}& 81,299& 81,299& 162,598& 17& GAN + Diffusion
\\

\midrule

\multirow{4}{*}{\textbf{In-the-Wild}}

& Chameleon~\cite{DBLP:conf/iclr/YanLCHJ0X25}& 14,863& 11,170& 26,033& Unknown& Unknown
\\

& SynthWildX~\cite{DBLP:conf/cvpr/CozzolinoPCNV22}& 421& 1,318& 1,739& 3& Diffusion
\\

& WildRF~\cite{DBLP:journals/corr/abs-2406-09398}& 1,251& 1,252& 2,503& Unknown& Unknown
\\

& BFree-Online~\cite{DBLP:conf/cvpr/GuillaroZUSCV25}& 271& 542& 813& Unknown& Unknown
\\

\bottomrule

\end{tabular}
\caption{
Statistics of the nine evaluation benchmarks used in this work.
curated datasets contain images generated by known synthesis models,
whereas in-the-wild datasets are collected from real online scenarios.
``Generators'' denotes the number of distinct image generators.
}
\label{tab:dataset_statistics}
\end{table*}

\section{Dataset Details}
\label{sec:supp_dataset}

\begin{table*}[!ht]
\centering
\scriptsize
\setlength{\tabcolsep}{2.5pt}

\resizebox{\textwidth}{!}{
\begin{tabular}{lccccccccccccccccc}

\toprule

Method
&LDM&SD1.4&SD1.5&SD2&SDXL&SDXL-R&SD-T&SDXL-T&LCM1.5&LCM-XL&SD1-C&SD2-C&SDXL-C&SD1-I&SD2-I&SDXL-I
&Avg.
\\

\midrule

NPR~\cite{DBLP:conf/cvpr/TanLZWGLW24}
&33.0&29.1&29.0&35.1&33.2&28.4&27.9&27.9&29.4&30.2&28.4&28.3&34.7&67.9&67.4&66.1&37.3\\

UnivFD~\cite{DBLP:conf/cvpr/OjhaLL23}
&85.4&56.8&56.4&58.2&63.2&55.0&56.5&53.0&54.5&65.9&68.0&65.4&75.9&64.6&56.2&53.9&61.8\\

FatFormer~\cite{DBLP:conf/cvpr/LiuTTW0Z24}
&55.9&48.2&48.2&48.2&48.2&48.3&48.2&48.2&48.3&50.6&49.7&49.9&59.8&66.3&60.6&56.0&52.2\\

SAFE~\cite{DBLP:conf/kdd/LiCHJHF25}
&50.3&50.1&50.0&50.0&49.9&50.1&50.0&50.0&50.1&50.0&49.9&50.0&54.7&98.2&98.5&97.3&59.3\\

C2P-CLIP~\cite{DBLP:conf/aaai/TanTLGWZW25}
&83.0&51.7&51.7&52.9&51.9&64.6&51.7&50.6&52.0&66.1&56.9&54.7&77.8&67.2&57.1&56.7&59.2\\

AIDE~\cite{DBLP:conf/iclr/YanLCHJ0X25}
&64.4&74.9&75.1&58.5&53.5&66.3&52.8&52.8&70.0&54.3&65.9&53.6&53.9&95.3&73.3&69.0&64.6\\

DRCT~\cite{DBLP:conf/icml/ChenZYY24}
&96.7&96.3&96.3&94.9&96.2&93.5&93.4&92.9&91.2&95.0&95.6&92.7&92.0&94.1&69.6&57.4&90.5\\

AlignedForensics~\cite{DBLP:conf/iclr/RajanOSL25}
&\textbf{99.9}&\textbf{99.9}&\textbf{99.9}&\textbf{99.6}&90.2&81.3&\textbf{99.7}&89.4&\textbf{99.7}&90.0&\textbf{99.9}&\underline{99.2}&87.6&\textbf{99.9}&\textbf{99.8}&92.6&95.5\\

DDA~\cite{DBLP:conf/nips/ChenXYZWXCXGYD25}
&99.2&98.9&\underline{99.0}&98.3&\underline{98.0}&\underline{96.8}&97.9&\underline{94.8}&95.9&\underline{98.2}&98.7&99.0&\underline{99.4}&99.0&\underline{99.5}&\underline{96.3}&\underline{98.1}\\

\midrule


PatchHead(ours)
&\underline{99.4}
&\underline{99.1}
&\underline{99.0}
&\underline{98.9}
&\textbf{99.1}
&\textbf{99.2}
&\underline{99.2}
&\textbf{98.8}
&\underline{99.3}
&\textbf{99.3}
&\underline{99.4}
&\textbf{99.4}
&\textbf{99.5}
&\underline{99.4}
&99.4
&\textbf{98.5}
&\textbf{99.2}
\\

\bottomrule

\end{tabular}
}
\caption{
Balanced accuracy (\%) on DRCT-2M.
The best and second-best results for each generator are highlighted in \textbf{bold} and \underline{underline}. * indicates values corrected from the original reports (DDA~\cite{DBLP:conf/nips/ChenXYZWXCXGYD25}).
}
\label{tab:drct2m}
\end{table*}

\begin{table*}[!ht]
\centering
\scriptsize
\fontsize{7}{8}\selectfont
\setlength{\tabcolsep}{1.5pt}

\resizebox{\textwidth}{!}{
\begin{tabular}{lcccccccccc}
\toprule

Method
& DALL$\cdot$E2
& DALL$\cdot$E3
& Firefly
& GLIDE
& Midjourney
& SD1.3
& SD1.4
& SD2
& SDXL
& Avg.
\\

\midrule

NPR~\cite{DBLP:conf/cvpr/TanLZWGLW24}
&51.1&49.3&46.5&48.5&52.8&51.4&51.8&46.0&52.8&50.0\\

UnivFD~\cite{DBLP:conf/cvpr/OjhaLL23}
&83.5&47.4&89.9&53.3&52.5&70.4&69.9&75.7&68.0&67.8\\

FatFormer~\cite{DBLP:conf/cvpr/LiuTTW0Z24}
&59.4&39.5&60.3&72.7&44.4&53.7&54.0&52.3&69.1&56.1\\

SAFE~\cite{DBLP:conf/kdd/LiCHJHF25}
&58.0&9.9&10.3&52.2&56.7&59.4&59.1&53.0&59.5&46.5\\

C2P-CLIP~\cite{DBLP:conf/aaai/TanTLGWZW25}
&55.6&63.2&59.5&86.7&52.9&75.2&76.7&69.2&77.7&68.5\\

AIDE~\cite{DBLP:conf/iclr/YanLCHJ0X25}
&34.9&33.7&24.8&65.0&57.5&74.1&73.7&53.2&68.4&53.9\\

DRCT~\cite{DBLP:conf/icml/ChenZYY24}
&77.2&86.6&84.1&82.6&73.7&86.6&86.6&83.2&71.3&81.3\rlap{$^*$}\\

AlignedForensics~\cite{DBLP:conf/iclr/RajanOSL25}
&50.2&48.9&51.7&53.5&\textbf{98.7}&\textbf{98.8}&\textbf{98.8}&\textbf{98.6}&\textbf{97.3}&77.4\\

DDA~\cite{DBLP:conf/nips/ChenXYZWXCXGYD25}
&\underline{86.3}&\underline{90.0}&\underline{91.9}&\underline{76.5}&\underline{93.5}&\underline{92.9}&\underline{92.7}&\underline{93.3}&\underline{93.5}&\underline{90.1}\\

\midrule


PatchHead(ours)
&\textbf{92.1}
&\textbf{92.4}
&\textbf{92.3}
&\textbf{90.1}
&92.5
&92.5
&92.5
&92.5
&92.5
&\textbf{92.1}
\\

\bottomrule
\end{tabular}
}
\caption{
Balanced Accuracy (\%) on the Synthbuster benchmark.
}
\label{tab:synthbuster}
\end{table*}

\begin{table}[!ht]
\centering
\scriptsize
\fontsize{7}{8}\selectfont
\setlength{\tabcolsep}{1.5pt}

\resizebox{\columnwidth}{!}{
\begin{tabular}{lcccccccc}
\toprule

\multirow{2}{*}{Method}
&
\multirow{2}{*}{Real}
&
\multicolumn{6}{c}{\textbf{Fake}}
&
\multirow{2}{*}{Avg.}
\\

\cmidrule(lr){3-8}

&
&XL
&EMA
&MSE
&SD2.1
&SD3.5
&FLUX.1
&
\\

\midrule

NPR~\cite{DBLP:conf/cvpr/TanLZWGLW24}
&55.4&16.1&31.1&41.3&41.2&24.9&19.2&42.2
\\

UnivFD~\cite{DBLP:conf/cvpr/OjhaLL23}
&99.2&3.9&9.6&7.3&7.4&3.4&1.8&52.4
\\

FatFormer~\cite{DBLP:conf/cvpr/LiuTTW0Z24}
&96.4&5.4&6.9&10.4&10.3&6.6&2.8&51.7
\\

SAFE~\cite{DBLP:conf/kdd/LiCHJHF25}
&98.8&0.6&0.9&0.9&1.0&0.3&1.8&49.9
\\

C2P-CLIP~\cite{DBLP:conf/aaai/TanTLGWZW25}
&99.5&2.0&2.7&4.3&4.2&4.0&1.3&51.3
\\

AIDE~\cite{DBLP:conf/iclr/YanLCHJ0X25}
&98.8&0.6&2.2&1.7&1.8&0.3&0.6&50.0
\\

DRCT~\cite{DBLP:conf/icml/ChenZYY24}
&94.2&16.9&34.8&33.5&33.6&21.7&17.2&60.2
\\

AlignedForensics~\cite{DBLP:conf/iclr/RajanOSL25}
&\textbf{99.8}&82.5&\underline{99.2}&99.0&99.1&55.4&3.6&86.5
\\

DDA~\cite{DBLP:conf/nips/ChenXYZWXCXGYD25}
&\underline{99.0}&\textbf{95.0}&\textbf{99.3}&\textbf{99.7}
&\textbf{99.7}&\underline{68.1}&\underline{50.2}&\underline{92.2}
\\

\midrule

PatchHead (ours)
&98.9&\underline{94.4}&\underline{99.2}&\underline{99.4}&\underline{99.4}
&\textbf{75.2}&\textbf{80.3}&\textbf{95.1}
\\

\bottomrule

\end{tabular}
}
\caption{
Balanced Accuracy (\%) on DDA-COCO.
}
\label{tab:ddacoco}
\end{table}

\begin{table}[!ht]
\centering
\scriptsize
\fontsize{7}{8}\selectfont
\setlength{\tabcolsep}{1.5pt}

\resizebox{\columnwidth}{!}{
\begin{tabular}{lcccccc}
\toprule

Method
&Flux
&GoT
&Infinity
&NOVA
&OmniGen
&Avg.
\\

\midrule

NPR~\cite{DBLP:conf/cvpr/TanLZWGLW24}
&0.7&0.2&6.5&4.7&2.2&2.9
\\

UnivFD~\cite{DBLP:conf/cvpr/OjhaLL23}
&4.0&9.2&15.7&8.3&39.6&15.4
\\

FatFormer~\cite{DBLP:conf/cvpr/LiuTTW0Z24}
&9.9&47.9&44.7&98.3&27.3&45.6
\\

SAFE~\cite{DBLP:conf/kdd/LiCHJHF25}
&1.0&0.5&1.9&0.6&1.6&1.1
\\

C2P-CLIP~\cite{DBLP:conf/aaai/TanTLGWZW25}
&8.7&49.6&35.3&86.4&14.5&38.9
\\

AIDE~\cite{DBLP:conf/iclr/YanLCHJ0X25}
&17.9&24.7&3.4&16.3&33.4&19.1
\\

DRCT~\cite{DBLP:conf/icml/ChenZYY24}
&72.5&81.4&77.9&84.6&72.5&77.8
\\

AlignedForensics~\cite{DBLP:conf/iclr/RajanOSL25}
&32.0&72.3&74.0&84.8&77.0&68.0
\\

DDA~\cite{DBLP:conf/nips/ChenXYZWXCXGYD25}
&\underline{89.9}&\underline{99.5}&\underline{97.8}&\underline{99.5}&\underline{99.5}&\underline{97.2}
\\

\midrule

PatchHead (ours)
&\textbf{97.5}
&\textbf{99.9}
&\textbf{100.0}
&\textbf{100.0}
&\textbf{99.9}
&\textbf{99.4}
\\

\bottomrule

\end{tabular}
}
\caption{
Balanced Accuracy (\%) on EvalGEN.
}
\label{tab:evalgen}
\end{table}

This section provides additional details and statistics of the nine
AI-generated image detection datasets used in the main experiments.
These datasets cover both manually curated benchmarks and
in-the-wild scenarios, representing diverse generation paradigms and
real-world distributions. Detailed statistics of all datasets are
summarized in Table~\ref{tab:dataset_statistics}.

The Manually Curated Datasets include DRCT-2M~\cite{DBLP:conf/icml/ChenZYY24},
DDA-COCO~\cite{DBLP:conf/nips/ChenXYZWXCXGYD25},
EvalGEN~\cite{DBLP:conf/nips/ChenXYZWXCXGYD25},
Synthbuster~\cite{synthbuster}, and
AIGCDetectionBenchmark~\cite{zhong2023patchcraft}.
These datasets contain images generated by known generative models and cover diverse generation paradigms, including GANs, diffusion models, flow matching models, and autoregressive models. They provide controlled benchmarks for measuring cross-generator generalization.

The In-the-Wild Datasets include Chameleon~\cite{DBLP:conf/iclr/YanLCHJ0X25},
SynthWildX~\cite{DBLP:conf/cvpr/CozzolinoPCNV22},
WildRF~\cite{DBLP:journals/corr/abs-2406-09398}, and
BFree-Online~\cite{DBLP:conf/cvpr/GuillaroZUSCV25}.
These datasets are collected from real-world online environments and contain more complex data distributions and unknown generation processes, providing challenging scenarios for evaluating the robustness of AI-generated image detectors.

\section{Implementation Details}
\label{sec:supp_protocol}
The training data is constructed from the DDA-aligned COCO-SD-2 dataset~\cite{DBLP:conf/nips/ChenXYZWXCXGYD25}, which contains real images from MS COCO and corresponding VAE reconstructed images. We adopt DINOv3 ViT-L/16 as the visual encoder and employ LoRA for
parameter-efficient fine-tuning. Specifically, LoRA modules are inserted into the combined QKV projection, attention output projection, and two MLP linear layers of each Transformer block, while the original backbone weights remain frozen.

To improve the robustness of the model against diverse generation
artifacts and real-world image degradations, we employ multiple data
augmentation strategies during training, including pixel-domain mixing,
frequency-domain mixing, and random JPEG compression.
Furthermore, to construct augmented views required for contrastive
learning, each training image undergoes an additional scale
transformation, where the image is first randomly downsampled or upsampled and then resized back to the original input size to generate another view.

The model is optimized using the AdamW optimizer with an initial learning
rate of $1\times10^{-4}$.
A cosine annealing strategy is adopted for learning rate scheduling.
The training batch size is set to 16 with 4 gradient accumulation steps,
resulting in an effective batch size of 64.
All experiments are conducted
on a single NVIDIA GeForce RTX 5090 GPU.

Unless otherwise specified, the weight of the auxiliary contrastive learning loss is fixed to 0.5.
All experiments are conducted under a single-model evaluation setting,
without dataset-specific fine-tuning or test-time adaptation.

\section{Detailed Cross-Dataset Evaluation}
\label{sec:supp_results}

We provide detailed cross-dataset results on more benchmarks in Tables~\ref{tab:drct2m}-\ref{tab:evalgen}, including DRCT-2M, Synthbuster, DDA-COCO, and EvalGEN. 
These results further validate the effectiveness and generalization ability of \method{} across diverse generation sources, covering both large-scale benchmarks.

On DRCT-2M, \method{} achieves the best average balanced accuracy of $99.2\%$, outperforming existing approaches and maintaining consistently strong performance across 16 generation sources.
On Synthbuster, \method{} reaches an average balanced accuracy of $92.1\%$, achieving the best results across most evaluated generators. Notably, it maintains a balanced accuracy above $90.1\%$ even on the most challenging generation source, improving by $13.6$ points over the second-best method.
For DDA-COCO and EvalGEN, \method{} achieves average balanced
accuracies of $95.1\%$ and $99.4\%$, respectively, outperforming
previous best results by $2.9$ and $2.2$ points. Overall, these results demonstrate that \method{} generalizes effectively across diverse manually curated benchmarks.

\section{Ablation Study Details}
\label{sec:supp_ablation}

\subsection{Analysis of Spatial Responses and Visualization }
\label{sec:supp_visualization}
This section provides additional details for the
\emph{Authenticity-Adaptive Spatial Responses} analysis in the main paper. We describe the visualization methods used to generate the spatial response maps and provide detailed definitions of the quantitative metrics used to characterize spatial responses.

\begin{figure*}[t]
    \centering
    \includegraphics[
        width=\textwidth
    ]{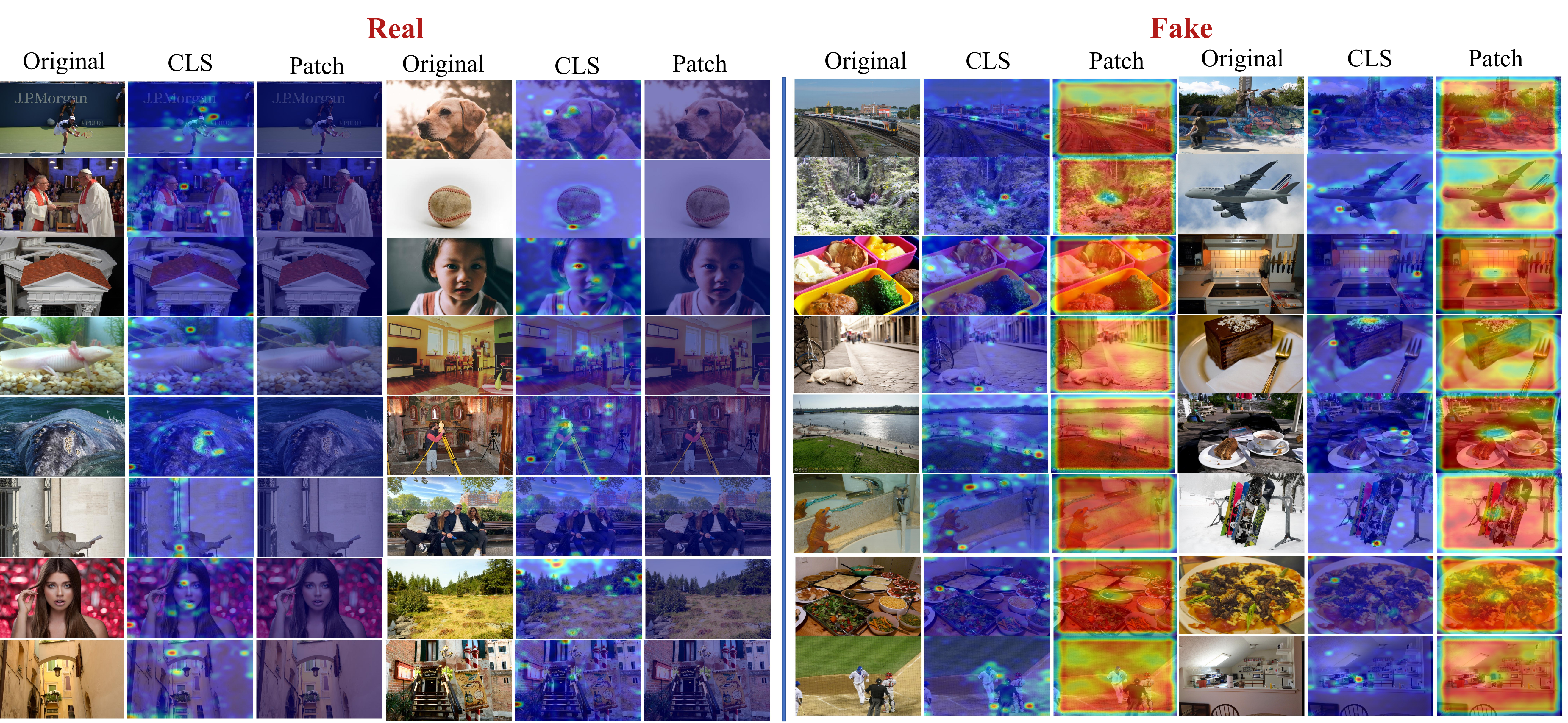}
    \caption{
    Additional qualitative visualization of spatial responses.
    }
    \label{fig:additional_visualization}
\end{figure*}

We employ different attribution methods for CLS representations and PatchHead, respectively. Specifically, Attention Rollout is adopted for CLS-based detectors to estimate patch-level contributions from transformer attention maps, while Class Activation Mapping (CAM) is used for PatchHead to generate
region-level response maps from the patch-based classification head. These attribution methods are selected according to the corresponding prediction mechanisms of the two architectures.
Additional visualization examples are provided in
Figure~\ref{fig:additional_visualization} to further illustrate the spatial response patterns across different samples.

For the quantitative analysis, we introduce four complementary metrics to
characterize spatial responses: Coverage, Top-10 Activation Ratio, Coverage
Histogram, and Spatial Variance. Coverage measures the spatial extent of
activated regions, Top-10 Activation Ratio evaluates the concentration of
responses on the most activated patches, Coverage Histogram describes the
distribution of spatial response ranges across samples, and Spatial
Variance measures the structural variation of response maps.

Although the two attribution methods are derived from different
prediction mechanisms, both produce spatial response maps indicating
prediction-related regions. Before quantitative analysis, all response
maps are normalized using the same Min-Max normalization procedure,
making the subsequent spatial statistics comparable across methods.

\begin{figure*}[!ht]
    \centering

    \begin{minipage}{0.48\textwidth}
        \centering
        \includegraphics[
            width=\linewidth
        ]{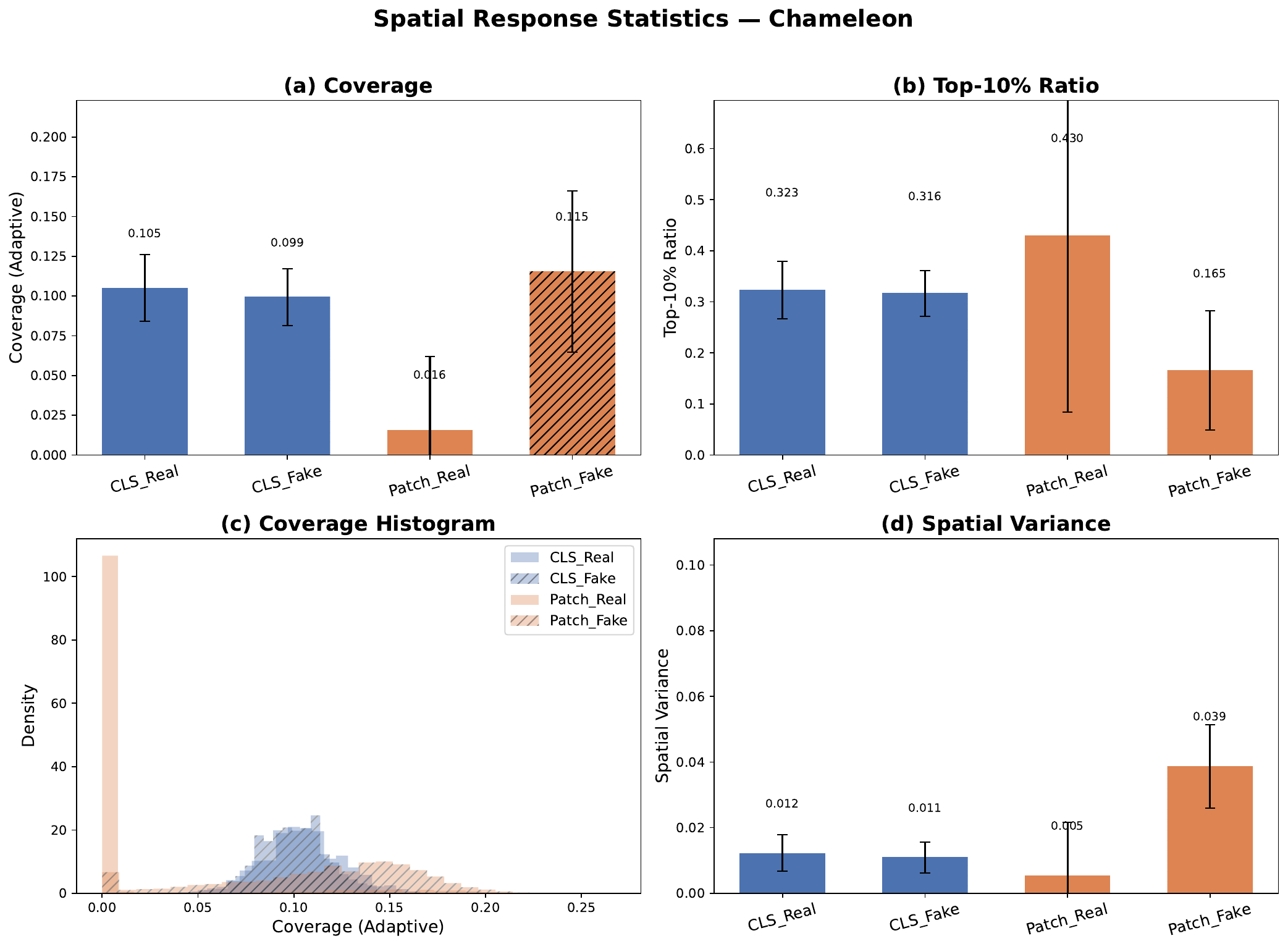}
    \end{minipage}
    \hfill
    \begin{minipage}{0.48\textwidth}
        \centering
        \includegraphics[
            width=\linewidth
        ]{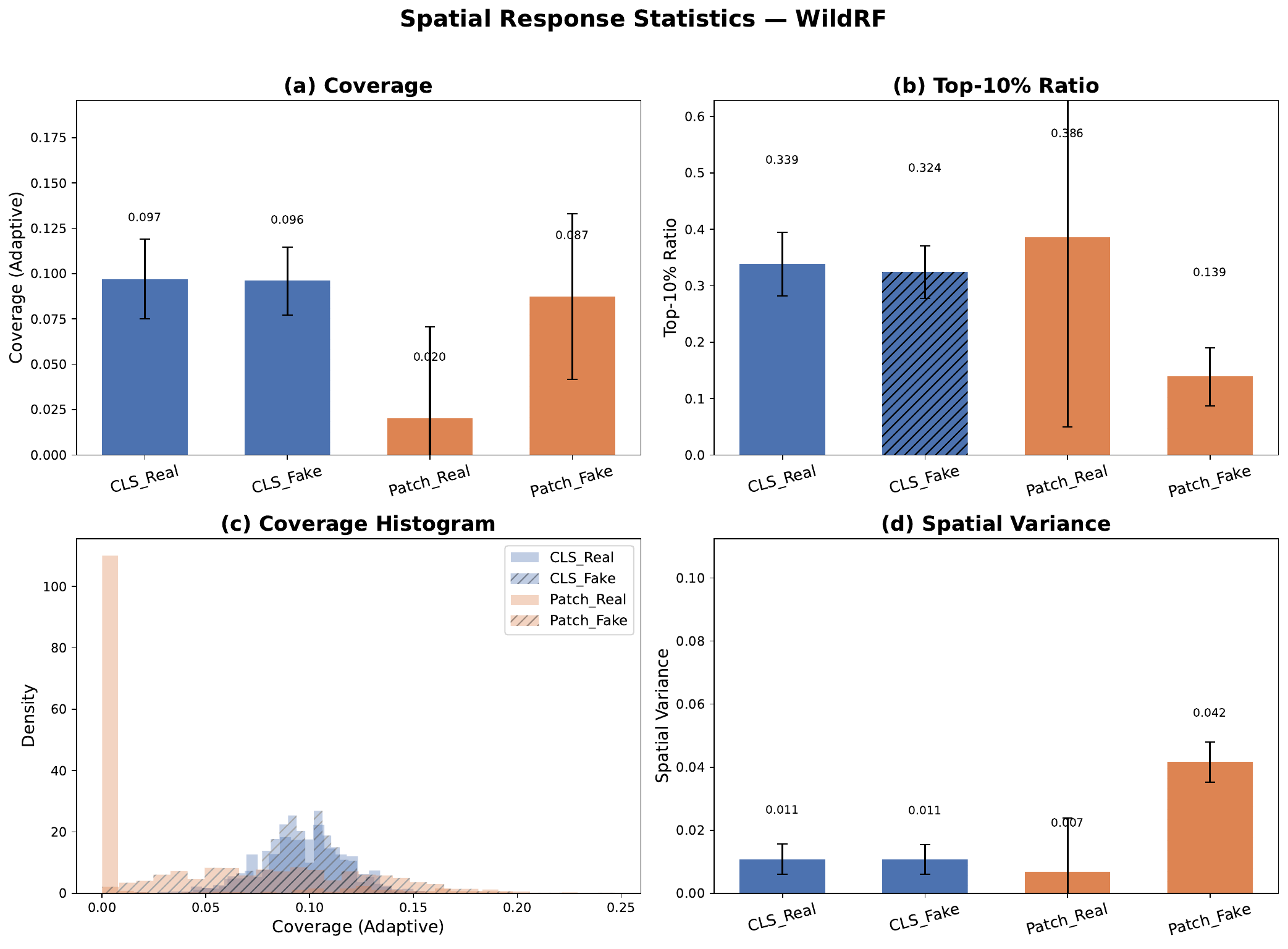}
    \end{minipage}

    \caption{
    Additional quantitative spatial response analysis on Chameleon and WildRF datasets.
    }
    \label{fig:additional_spatial_analysis}
\end{figure*}

In addition to the BFree-Online results reported in the main paper, we
further provide quantitative results on Chameleon and WildRF in
Figure~\ref{fig:additional_spatial_analysis} to verify the consistency of
the observed spatial response patterns across different datasets.

\paragraph{Coverage}

Coverage measures the proportion of spatial regions where the model
produces significant responses. Given the normalized response map
$\hat{H}\in\mathbb{R}^{H\times W}$, Coverage is defined as:

\begin{equation}
Coverage=\frac{1}{HW}\sum_{i,j}\mathbb{I}(\hat{H}_{ij}>\tau),
\end{equation}

where $\tau$ denotes the response threshold. In our experiments,
$\tau$ is adaptively determined as the mean value plus one standard
deviation of each response map:

\begin{equation}
\tau=\mu_{\hat{H}}+\sigma_{\hat{H}}.
\end{equation}

This adaptive criterion avoids manually selected fixed thresholds and
provides a sample-specific measurement of significant spatial activation.
A larger Coverage indicates that the model activates a broader spatial
range.

As shown in Figure~\ref{fig:additional_spatial_analysis}(a), CLS
representations exhibit similar Coverage distributions between real and
generated images on both datasets. In contrast, PatchHead maintains low
Coverage on real images while producing substantially higher Coverage on
generated images. This indicates that PatchHead produces
category-dependent spatial responses, with broader activations on
generated images.

\paragraph{Top-10 Activation Ratio}

To measure the concentration of spatial responses, we introduce the
Top-10 Activation Ratio. Given the flattened normalized response vector
$v=[v_1,\cdots,v_N]$ sorted in descending order, the Top-10 Activation
Ratio is defined as:

\begin{equation}
Top10=\frac{\sum_{i=1}^{K}v_i}{\sum_{i=1}^{N}v_i},
\end{equation}

where $K=0.1N$.

A higher Top-10 Activation Ratio indicates that a small subset of patches contributes most of the overall response, whereas a lower ratio indicates that the final decision is supported by more distributed spatial evidence.

Figure~\ref{fig:additional_spatial_analysis}(b) shows that CLS representations maintain relatively high Top-10 Activation Ratios for both real and generated images. By contrast, PatchHead achieves lower Top-10 Activation Ratios on generated images, suggesting that it
aggregates authenticity-related evidence from multiple spatial patches rather than relying on a few dominant regions.

\paragraph{Coverage Histogram}

Since averaged statistics may conceal sample-level variations, we further analyze the complete Coverage distribution over all samples.

As shown in Figure~\ref{fig:additional_spatial_analysis}(c), the Coverage distributions of CLS representations for real and generated images are highly overlapped. In contrast, PatchHead exhibits a clear distribution shift: real images are concentrated in the low-Coverage region, while generated images show consistently higher Coverage values. This indicates that the observed spatial response difference consistently appears across samples rather than being dominated by individual cases.

\paragraph{Spatial Variance}

Finally, we introduce Spatial Variance to measure the structural
difference of response maps.

It is defined as:

\begin{equation}
Var=\frac{1}{N}\sum_{i=1}^{N}(\hat{H}_i-\mu_{\hat H})^2,
\end{equation}

where $\mu$ represents the mean value of the response map. A larger
Spatial Variance indicates stronger spatial variation and more
distinctive response structures.

Figure~\ref{fig:additional_spatial_analysis}(d) shows that CLS
representations exhibit limited Spatial Variance differences between real and generated images, suggesting similar response structures across categories. In contrast, PatchHead maintains lower variance on real images while producing substantially higher variance on generated images. This indicates that PatchHead produces more diverse spatial response structures for generated images.

\subsection{Architecture of Complex Spatial Designs}

\label{sec:supp_design}
This section provides detailed descriptions of three variants based on the default PatchHead architecture, as introduced in the main paper. All variants use the same DINO backbone and training objective as PatchHead, while introducing additional spatial modeling components from three perspectives: multi-scale receptive field aggregation, multi-level feature fusion, and HRM-inspired iterative refinement. The detailed architectures are described as follows.

\paragraph{Multi-scale}

The Multi-scale variant extends PatchHead by incorporating multiple spatial receptive fields. Given the final-layer patch tokens, the features are first projected from the backbone dimension into the PatchHead feature dimension and reshaped into a two-dimensional spatial feature map.

Three parallel branches with different receptive fields are then applied: a $1\times1$ convolution branch for channel-wise feature interaction, a $3\times3$ depth-wise convolution branch for local spatial aggregation, and a dilated $3\times3$ depth-wise convolution branch for capturing larger spatial context. The outputs from different branches are adaptively fused using image-dependent weights.

The fused features are aggregated using a statistical pooling strategy that combines global feature statistics (mean and standard deviation) with a soft-weighted aggregation of Top-K high-response patches. The resulting representation is then fed into the classification head for final prediction.

\paragraph{Multi-level}

The Multi-level variant incorporates patch features from multiple DINO layers. Specifically, patch representations from layers 8, 12, 16, and 23 are extracted and projected into a shared feature space.

The features from different layers are fused as:

\begin{equation}
F_{\mathrm{ml}}
=
\sum_{l\in\{8,12,16,23\}}
\alpha_l \phi(X_l),
\end{equation}

where $X_l$ denotes the patch representation from the $l$-th DINO layer, $\phi(\cdot)$ represents the shared feature projection, and $\alpha_l$ denotes the learnable weight of each layer.

The fused representation is further combined with the final-layer feature through a learnable residual scaling operation and then reshaped into a spatial feature map. The resulting feature is processed by the spatial aggregation module used in the Multi-scale variant.

\paragraph{HRM-inspired}

The HRM-inspired variant introduces an iterative refinement mechanism inspired by the Hierarchical Reasoning Model (HRM)~\cite{wang2025hierarchical}. Instead of directly aggregating patch features in a single stage, this variant maintains two interacting latent states: a refinement state for progressively updating spatial information and an output state for generating the final prediction representation.

At each refinement step, the two states are updated through interactions with patch features and each other:

\begin{equation}
r_{t+1}=f_r(r_t,a_t,X),
\end{equation}

\begin{equation}
a_{t+1}=f_a(a_t,r_{t+1},X),
\end{equation}

where $r_t$ and $a_t$ denote the two latent states at step $t$,
respectively, and $X$ represents the patch features.

The refinement process is performed iteratively with shared parameters.
The final prediction is obtained from the refined output state through a gated pooling mechanism that aggregates spatial statistics.

\begin{table*}[!t]
\centering
\resizebox{\textwidth}{!}{
\begin{tabular}{lcccccccccc}
\toprule
Configuration &
Cham. &
AIGCBench &
DRCT-2M &
SynthBuster &
SynthWildX &
WildRF &
BFree &
DDA-COCO &
EvalGEN &
Avg. \\
\midrule

Rank=8, $\alpha$=1(default)
& \underline{92.0} & \textbf{96.2} & \underline{99.2} & 92.1 & 89.4 & \underline{93.4} & \textbf{94.6} & 95.1 & \textbf{99.4} & \underline{94.6} \\

Rank=8, $\alpha$=2
& 91.0 & 95.5 & 99.1 & \underline{94.5} & 89.2 & 92.1 & 93.6 & \textbf{95.9} & 98.9 & 94.0 \\

Rank=8, $\alpha$=4
& 91.9 & 95.3 & \textbf{99.4} & \textbf{95.4} & \textbf{89.5} & 92.1 & \underline{94.2} & 95.5 & 99.1 & \textbf{94.7} \\

Rank=16, $\alpha$=1
& \textbf{92.2} & \underline{96.0} & \underline{99.2} & 91.9 & 89.4 & 93.3 & 92.2 & \underline{95.6} & 99.1 & 94.3 \\

Rank=32, $\alpha$=1
& 91.6 & 95.7 & \underline{99.2} & 91.7 & 89.4 & \textbf{94.2} & 94.5 & 95.2 & \underline{99.4} & 94.5 \\
\bottomrule
\end{tabular}
}
\caption{
Ablation study of different LoRA configurations on nine cross-dataset benchmarks.
Results are reported in Balanced Accuracy (\%). Best results are shown in \textbf{bold}, and second-best results are \underline{underlined}.
}
\label{tab:lora_ablation}
\end{table*}

\subsection{Ablation Study on LoRA Configuration}
\label{sec:supp_lora}
To analyze the sensitivity to LoRA configurations, we evaluate different combinations of rank and scaling factor $\alpha$ on nine cross-dataset benchmarks. As shown in Table~\ref{tab:lora_ablation}, different LoRA settings result in relatively small performance variations. Increasing the rank or $\alpha$ value may improve individual datasets, but these gains are not consistently observed across different domains. For example, larger $\alpha$ values improve performance on SynthBuster and DDA-COCO, while slightly degrading results on other benchmarks, indicating that increasing LoRA capacity within the evaluated range does not consistently improve cross-domain generalization.

We adopt Rank=8 with $\alpha=1$ as the default configuration. Although Rank=8 with $\alpha=4$ achieves a slightly higher average score
(by 0.1 percentage points), the improvement is marginal, and the default configuration provides more balanced performance across different datasets.

\end{document}